\def\year{2019}\relax
\documentclass[letterpaper]{article} 
\usepackage{aaai19}  
\usepackage{times}  
\usepackage{helvet}  
\usepackage{courier}  
\usepackage{url}  
\usepackage{graphicx}  
\usepackage{amsmath,amssymb} 
 
 \usepackage{subcaption}
 \usepackage{multirow}
 \usepackage{booktabs}
 \usepackage{amsmath}

 \usepackage{xcolor}
 
\begin{document}
%
\title{Deep Metric Learning by Online Soft Mining and Class-Aware Attention}

\author{Xinshao Wang\textsuperscript{1,2}, Yang Hua\textsuperscript{1}, Elyor Kodirov\textsuperscript{2}, Guosheng Hu\textsuperscript{2,1}, Neil M. Robertson\textsuperscript{1,2} \\
\textsuperscript{1} School of Electronics, Electrical Engineering and Computer Science, Queen's University Belfast, UK \\
\textsuperscript{2} Anyvision Research Team, UK\\
\{xwang39, y.hua, n.robertson\}@qub.ac.uk, \{elyor, guosheng.hu\}@anyvision.co
}

\maketitle

\begin{abstract}
	Deep metric learning aims to learn a deep embedding that can capture the semantic similarity of data points. Given the availability of massive training samples, deep metric learning is known to suffer from slow convergence due to a large fraction of trivial samples. Therefore, most existing methods generally resort to sample mining strategies for selecting nontrivial samples to accelerate convergence and improve performance. In this work, we identify two critical limitations of the sample mining methods, and provide solutions for both of them. 
	First, previous mining methods assign one binary score to each sample, i.e., dropping or keeping it, so they only selects a subset of relevant samples in a mini-batch. Therefore, we propose a novel sample mining method, called Online Soft Mining (OSM), which assigns one continuous score to each sample to make use of all samples in the mini-batch. OSM learns extended manifolds that preserve useful intraclass variances by focusing on more similar positives. 
	Second, the existing methods are easily influenced by outliers as they are generally included in the mined subset. 
	To address this, we introduce Class-Aware Attention (CAA) that assigns little attention to abnormal data samples. Furthermore, by combining OSM and CAA, we propose a novel weighted contrastive loss to learn discriminative embeddings. 
	Extensive experiments on two fine-grained visual categorisation datasets and two video-based person re-identification benchmarks show that our method significantly outperforms the state-of-the-art.
	Our source code is available at: {\color{blue} \url{https://github.com/XinshaoAmosWang/OSM_CAA_WeightedContrastiveLoss}}.
\end{abstract}

\begin{figure*}[t]
	\centering
	\includegraphics[width=\linewidth]{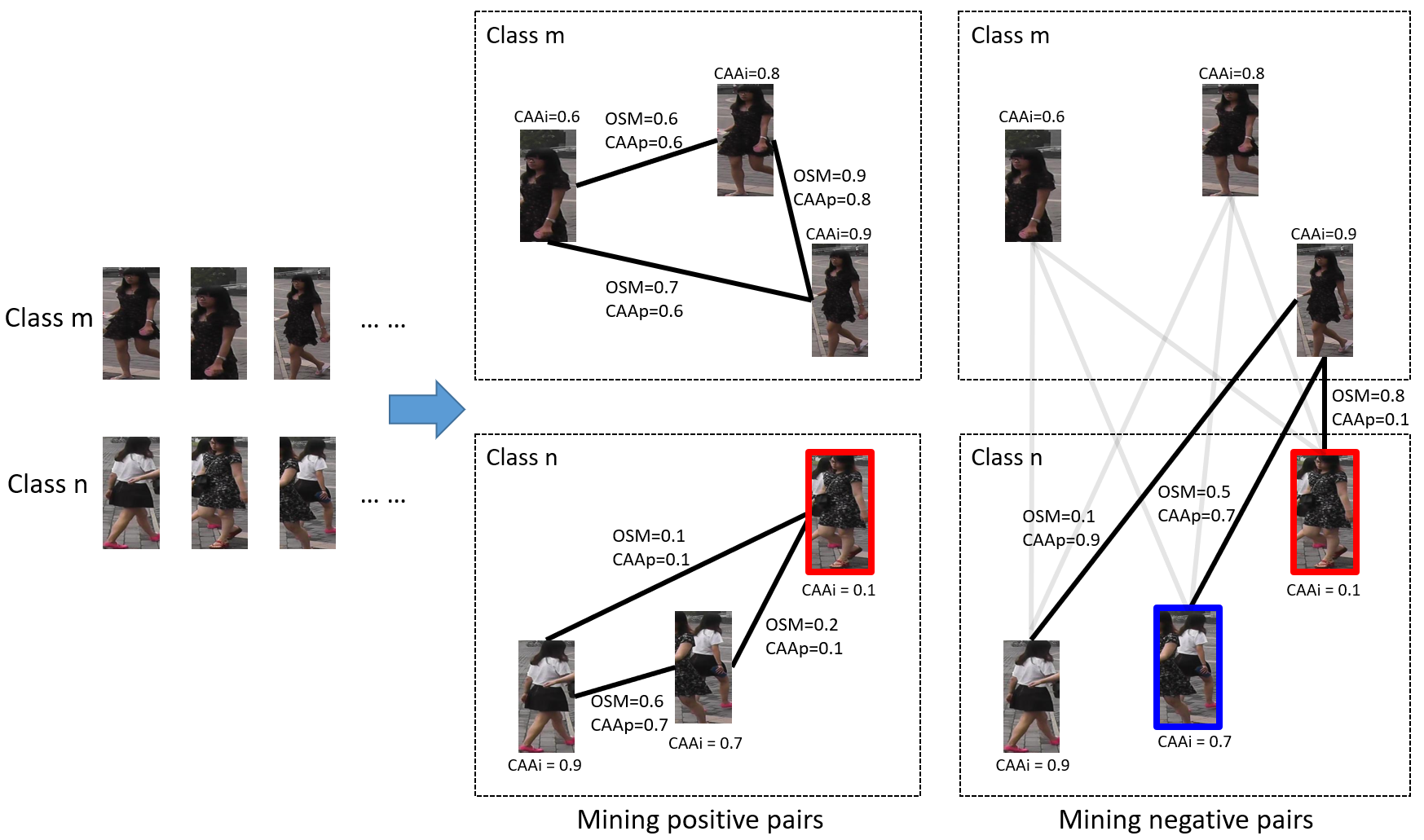}
	\caption{
		The illustration of Online Soft Mining (OSM) and Class-Aware Attention (CAA) for pair mining. For simplicity, we only consider the scenario with class $m$ and class $n$. For mining positive pairs, OSM focuses on local positives, i.e., assigning higher scores to the pairs with smaller distance. 
		Meanwhile, CAA helps identify outlier (with red boundary) by CAA score of images ($CAA_i$) and pairs ($CAA_p$). 
		For mining negative pairs (only highlighting pairs of one image from class $m$), OSM generates higher mining scores for more difficult negatives, e.g., outlier (with red boundary) and true difficult negative (with blue boundary), then CAA helps ignore outliers in the difficult negatives mined by OSM. \textit{Best viewed in colour.}
	}
	\label{fig:OSM_CAA}
\end{figure*}

\section{Introduction}
\label{introduction}
With success of deep learning, deep metric learning has attracted a great deal of attention and been applied to a wide range of visual tasks such as image retrieval \cite{wang2014learning,huang2016unsupervised,song2016deep}, face recognition and verification \cite{schroff2015facenet,sun2014deep,taigman2014deepface}, and person re-identification \cite{mclaughlin2017video,song2017region}. 
The goal of deep metric learning is to learn a feature embedding/representation that captures the semantic similarity of data points in the embedding space such that samples belonging to the same classes are located closer, while samples belonging to different classes are far away. 
Various deep metric learning methods have been proposed: doublet-based methods with contrastive loss \cite{song2016deep,hadsell2006dimensionality,shi2016embedding,yuan2017hard}, triplet-based methods with triple loss \cite{schroff2015facenet,wang2014learning,chechik2010large,hoffer2015deep,cui2016fine}, and quadruplet-based methods with double-header hinge loss \cite{huang2016local}. 
%
All of them construct pairs (samples) by using several images and introduce relative constraints among these pairs.

Compared with classification task, deep metric learning methods deal with much more training samples since all possible combinations of images need to be considered, e.g., quadratically larger samples for contrastive loss and cubically larger samples for triplet loss. 
Consequently, this yields too slow convergence and poor local optima due to a large fraction of trivial samples, e.g., non-informative samples that can be easily verified and contribute zero to the loss function. 
To alleviate this, sample mining methods are commonly used to select nontrivial images to accelerate convergence and improve performance. Therefore, a variety of sample mining strategies have been studied recently \cite{schroff2015facenet,song2016deep,wang2015unsupervised,simo2015discriminative,huang2016local,yuan2017hard,shi2016embedding,cui2016fine}.  


Despite the popularity of exploiting sample mining in deep metric learning, there are two problems in the existing methods:

(1) \textit{Not making full use of all samples in the mini-batch} -- the first problem concerns with the use of samples in the mini-batch. When most methods perform sample mining, they select samples out of mini-batch, e.g., \emph{hard}
\footnote{In this paper, `hard' means `not soft' as we propose soft mining (continuous mining score) in contrast with `hard' mining (binary mining score). While in previous work, it means `difficult'.}
 difficult negatives mining ~\cite{song2016deep}. Specifically, they assign one binary score (0 or 1) to each sample in order to `drop it' or `keep it'. In this way, some samples which may be useful to some extent are removed and thus only a subset of samples are mined. Therefore they cannot make full use of information in each mini-batch. 
 
 (2) \textit{Outliers} -- the second problem arises from outliers. The existing mining methods do not consider outliers. However, the presence of noise in the training set hurts the model's performance significantly as shown in \cite{zhang2016understanding}. In addition, difficult negatives mining prioritizes samples with larger training loss. As a result, outliers are generally mined in the subset for training as their training losses are more likely to be large.  

In this work, we aim to address both problems. Specifically, 
for the first problem, we propose a novel sample mining method, called Online Soft Mining (OSM), which assigns \emph{one continuous score} to each sample to make use of all samples in the mini-batch. OSM is based on Euclidean distance in the embedding space. 
OSM incorporates online soft positives mining and online soft negatives mining. 
For the positives mining, inspired by \cite{cui2016fine}, we \emph{assign higher mining scores to more similar pairs} in terms of distance. This mining has a strong connection to manifold learning \cite{cui2016fine}, since it preserves helpful intraclass variances like pose and colour in the deep embedding space. 
For the negatives mining, \emph{more difficult negatives get higher mining scores} by soft negatives mining, which can be considered as \emph{a generalisation of hard difficult negatives mining} \cite{song2016deep,wang2015unsupervised,simo2015discriminative,huang2016local,yuan2017hard,cui2016fine}.  
For the second problem, we propose Class-Aware Attention (CAA) that \emph{pays less attention to outlying data samples}. CAA is simply based on class compatibility score.
The illustration of OSM and CAA is given in Figure~\ref{fig:OSM_CAA}. 
One attractive property of OSM and CAA is that they are generic, by which we mean they can be integrated to most existing deep metric learning methods \cite{hadsell2006dimensionality,chechik2010large} for mining samples and addressing outliers simultaneously. 
To demonstrate the effectiveness of OSM and CAA, we propose a novel loss, Weighted Contrastive Loss (WCL), by integrating them into traditional contrastive loss \cite{hadsell2006dimensionality}. The weight is assigned to each pair of images defined by OSM and CAA.

Compared to previous hard mining (0/1) in metric learning, soft mining  is proposed novelly in this work. Besides, for the first time CAA is proposed to address the influence of outliers, which is an inherent problem of mining methods. 
Extensive experiments are conducted on two tasks: fine-grained visual categorisation and video-based person re-identification. 
For the first one, we demonstrate our proposed method's performance by comparing with other metric learning methods on CUB-200-2011 \cite{wah2011caltech} and CARS196 \cite{krause20133d} datasets. 
For the second one, we use MARS \cite{zheng2016mars} and LPW \cite{song2017region} datasets, and show that our method surpasses state-of-the-art methods by a large margin.

\section{Related Work}
\label{related_work}

\paragraph{Sample mining in deep metric learning.}
With the popularity of deep learning \cite{krizhevsky2012imagenet,szegedy2015going,girshick2016region}, deep metric learning has been widely studied in many visual tasks and achieved promising results \cite{schroff2015facenet,song2016deep}. 
Compared with traditional metric learning in which hard-crafted features are widely used \cite{chechik2010large,weinberger2009distance}, deep metric learning learns feature embeddings directly from data points using deep convolutional neural networks (CNN). 
Although we can construct a large number of image pairs for deep metric learning, a large fraction of trivial pairs will contribute zero to the loss and gradient once the model reaches a reasonable performance. 
Thus it is intuitive to mine nontrivial pairs during training to achieve faster convergence and better performance.
As a result, sample mining has been widely explored and a range of mining methods have been proposed \cite{schroff2015facenet,wang2015unsupervised,simo2015discriminative,huang2016local,yuan2017hard,shi2016embedding,cui2016fine,song2016deep}. 
For mining negatives, mining difficult negatives is applied in \cite{wang2015unsupervised,simo2015discriminative,huang2016local,song2016deep}. 
Mining semi-difficult negatives is studied in \cite{schroff2015facenet,cui2016fine}. 
For mining positives, mining difficult positives is explored in \cite{simo2015discriminative,huang2016local,yuan2017hard}. 
Mining semi-difficult positives is explored in \cite{shi2016embedding}. 
Local positives mining (i.e., closer positives) is proposed for learning an extended manifold rather than a contracted hypersphere in \cite{cui2016fine}. 
All of them \emph{mine a subset of pairs hardly} by using binary score for each pair, i.e., dropping it or keeping it. Instead, our proposed OSM conducts \emph{soft mining using continuous score}, i.e., assigning different weights for different pairs. 
In addition, samples with larger training loss are prioritized when difficult pair mining is applied.
Therefore, \emph{outliers are generally mined and perturb the model training} due to their large losses. 
To address this, semi-difficult negatives mining \cite{schroff2015facenet,cui2016fine} and multilevel mining \cite{yuan2017hard} are applied to remove outliers.    
Instead, we propose CAA for assigning smaller weights to outliers in the mini-batch.

In this paper, following~\cite{sohn2016improved,song2016deep,tadmor2016learning}, we construct one image pair between every two images in the mini-batch. $N$-pair-mc loss~\cite{sohn2016improved} and Multibatch~\cite{tadmor2016learning} treat all the constructed pairs equally, i.e., without any pair mining. LiftedStruct~\cite{song2016deep} mines most difficult negatives and treat all positives equally. In contrast, we propose online soft mining for both positives and negatives. 
%
\\
\noindent
\textbf{Intraclass variance.} 
In fine-grained recognition \cite{song2016deep,wah2011caltech,krause20133d,cui2016fine} and person re-identification \cite{zheng2016mars,zhong2017re,mclaughlin2017video,song2017region}, intraclass distance could be larger than interclass distance, e.g., images from different categories could have similar colour and shape while the images in the same category can have large variances such as colour, pose and lighting. Several approaches benefit from utilising the inherent structure of data. In particular,
local similarity-aware embedding was proposed in \cite{huang2016local} to preserve local feature structures. \cite{cui2016fine} proposed to learn an extended manifold by using only local positives rather than treating all positives equally. In a similar spirit with \cite{cui2016fine}, our OSM assigns larger weights for more similar positives, intending to learn continuous manifolds. The key difference, however, is that a subset composed of local positives are mined in \cite{cui2016fine}, whereas our OSM makes use of all positives and gives local positives more attention.

\section{Methodology}
Given an image dataset $X = \{(\mathbf{x}_i, y_i)\}$, where  $\mathbf{x}_i$ and $y_i$ are $i$-th image and the corresponding label respectively, our aim is to learn an embedding function (metric) $f$ that takes $\mathbf{x}_i$ as input and outputs its embedding $\mathbf{f}_i \in \mathbb{R}^D$. 
Ideally, the learned features of similar pairs are pulled together and features of dissimilar pairs are pushed away.

\begin{figure*}
	\centering
	\begin{subfigure}[b]{0.345\textwidth}
		\includegraphics[width=\textwidth]{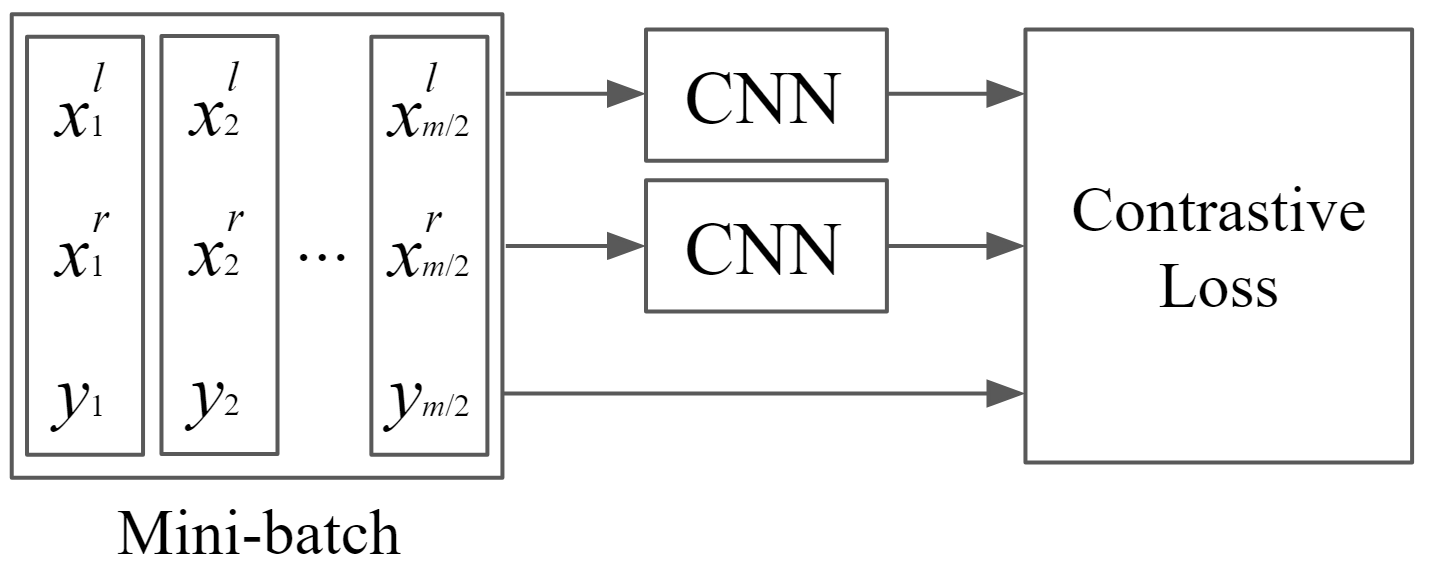}
		\caption{Traditional contrastive loss}
		\label{fig:traditional}
	\end{subfigure}
	\hfill
	~ 
	~ 
	\begin{subfigure}[b]{0.625\textwidth}
		\includegraphics[width=\textwidth]{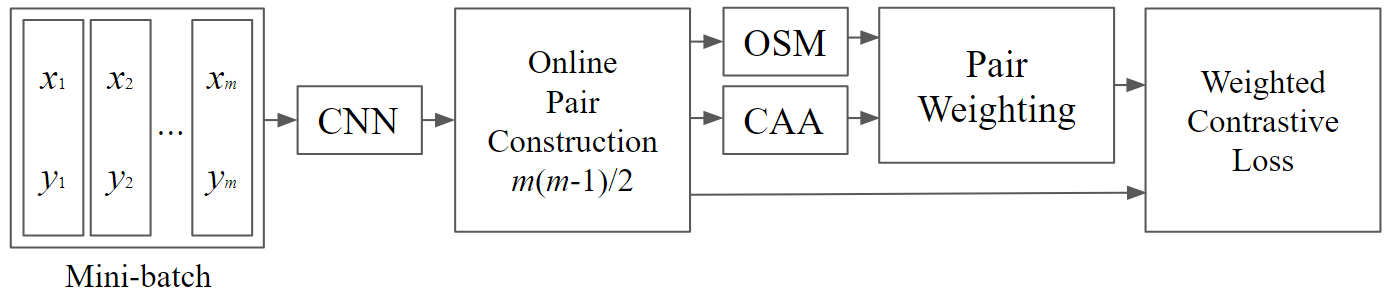}
		\caption{Weighted contrastive loss with OSM and CAA}
		\label{fig:weighted_contrastive_loss}
	\end{subfigure}
	\caption{
		The comparison of our proposed weighted contrastive loss with traditional contrastive loss. $m$ is the number of images in a mini-batch.  
		Traditional contrastive loss (a) takes as input $m/2$ image pairs and $m/2$ binary labels which indicate the corresponding image pair is from the same class or not.
		%
		Our proposed weighted contrastive loss (b) applies online pair construction. The weighted contrastive loss takes as input $m$ images and $m$ multi-class labels.
		We construct non-repeated $m(m-1)/2$ image pairs online rather than $m^2$ pairs in \cite{song2016deep} or $m(m-1)$ pairs in \cite{tadmor2016learning}.
		Weighted contrastive loss combines Online Soft Mining (OSM) and Class-Aware Attention (CAA) to assign a proper weight for each image pair. 
	}
	\label{fig:overview}
\end{figure*}

Our proposed weighted contrastive loss for learning the embedding function $f$ is illustrated in Figure \ref{fig:weighted_contrastive_loss}. 
Following the common practice, each mini-batch contains $m = c \times k$ images, where $c$ is the number of classes and $k$ is the number of images per class. 
CNN is used for extracting the feature of each image in the mini-batch. 
In online pair construction, we use all images from the same class to construct positive pairs (i.e., similar pairs) and all images from different classes to construct negative pairs (i.e., dissimilar pairs). 
As a result, we get $m(m-1)/2$ image pairs in total. There are $ck(k-1)/2$ positive pairs included in the positive set, i.e., $P = \{(\mathbf{x}_i, \mathbf{x}_j)|y_i = y_j\}$. Similarly, the negative set, $N = \{(\mathbf{x}_i, \mathbf{x}_j)|y_i \neq y_j\}$, consists of $ck(ck-k)/2$ negative pairs. 
%
Then, the weight of each pair is generated by OSM and CAA jointly. 
%
%
Our proposed weighted contrastive loss is based on these pairs and their corresponding weights. 
For reference, the traditional contrastive loss~\cite{hadsell2006dimensionality} is:
\begin{equation}
L^\alpha_{cont}(\mathbf{x}^{i}, \mathbf{x}^{j};f) = y_{ij}d_{ij}^2 + (1-y_{ij})max(0, \alpha - d_{ij})^2,
\end{equation}
where $y_{ij}=1$ if $y_{i}=y_{j}$, $y_{ij}=0$ if $y_{i} \neq y_{j}$, 
$d_{ij} = \left\Vert
\mathbf{f}_i - \mathbf{f}_j
\right\Vert_2 $  
is the euclidean distance between the pair. It pulls positive pairs as close as possible and pushes negative pairs farther than a pre-defined margin $\alpha$. 

%


\subsection{Online Soft Mining}
\label{sec:OSM}

Online Soft Mining contains Online Soft Positives Mining (OSPM) for the positive set and Online Soft Negatives Mining (OSNM) for the negative set. 
\\
\\
\noindent
\textbf{Online Soft Positives Mining.} The goal of OSPM is to generate OSM scores for pairs in the positive set. In many visual tasks, e.g., fine-grained visual categorisation, the interclass distance could be small compared to large intraclass distance. Motivated by manifold learning in \cite{cui2016fine} where \emph{local positives are selected out for learning extended manifolds}, OSPM assigns higher OSM scores to local positives. This is because contracted hyperspheres are learned and large intraclass distance cannot be captured if positives with large distance are treated equally.

Specifically, for each similar pair in the positive set, i.e., $(\mathbf{x}_i, \mathbf{x}_j) \in P$, we compute Euclidean distance $d_{ij}$ between their features after $L_2$ normalisation. 
As we want to assign higher mining scores to more similar pairs, we simply transfer the distance $d_{ij}$ to OSM score using a Gaussian function with mean=0. In summary, the OSM score $s^+_{ij}$ for each positive pair $(\mathbf{x}_i, \mathbf{x}_j)$ is obtained as follows:


\begin{equation}
	\label{equation:w+ij}
	s^+_{ij} = \exp
	({{{ - {d^2_{ij} }  } \mathord{\left/ {\vphantom {{ - \left( {s(\mathrm{x}_i) - \mu } \right)^2 } {2\sigma ^2 }}} \right. \kern-\nulldelimiterspace} { {\sigma_{\mbox {\tiny OSM}}} ^2 }}}
	)
\end{equation}
where $d_{ij} = \|\mathbf{f}_i - \mathbf{f}_j \|_2$, $\sigma_{\mbox {\tiny OSM}}$ is a hyperparameter for controlling the distribution of OSM scores.  
\\
\\
\noindent
\textbf{Online Soft Negatives Mining.} 
For dissimilar pairs in the negative set $N$, we wish to push each pair away by a margin $\alpha$ inspired by the traditional contrastive loss. 
Similar to previous hard difficult negatives mining methods using binary mining score \cite{song2016deep,yuan2017hard,cui2016fine,huang2016local,ustinova2016learning,tadmor2016learning}, to ignore a large fraction of trivial pairs which do not contribute to learning, we assign higher OSM scores to \emph{negative pairs whose distance is within this margin}. 
The OSM scores of negative pairs whose distances are larger than the pre-defined margin are set as 0 because \emph{they do not contribute to the loss and gradient}.  For simplicity, the OSM score $s^-_{ij}$ of each negative pair $(\mathbf{x}_i, \mathbf{x}_j)$ is computed directly by the margin distance: 
\begin{equation}
	\label{equation:w-ij}
	s^-_{ij} = \max(0, \alpha - d_{ij}) 
\end{equation}

\subsection{Class-Aware Attention}

As indicated in~\cite{schroff2015facenet,cui2016fine}, sample mining is easily influenced by outliers, leading the learned model to a bad local minimum. The proposed OSM is also prone to outliers, e.g., we assign more attention to more difficult negatives which may contain outliers. 

In general, outliers are generally caused by corrupted labels, which hurt the model's performance significantly \cite{zhang2016understanding}. In other words, outliers are usually composed of mislabelled images, thus being less semantically related to their labels. 
Therefore, we propose to identify noisy images by their semantic relation to their labels. As it is guided by the class label, we name it Class-Aware Attention (CAA). The CAA score of image $\mathbf{x}_i$ indicates how much it is semantically related to its label $y_i$.

To measure image's semantic relation to its label, we compute the compatibility between image's feature vector and its corresponding class context vector. The compatibility between two vectors can be measured by their dot product~\cite{jetley2018learn}. 
We apply a classification branch after image embeddings to learn context vectors of all classes. 
The class context vectors are the trained parameters of the fully connected layer, i.e., $\{ \mathbf{c}_k \}^{C}_{k=1}$, where $C$ is the number of classes in the training set and $\mathbf{c}_k \in \mathbb{R}^D$ is the context vector of class $k$. 
Therefore, the CAA score (image's semantic relation to label) of $\mathbf{x}_i$ is computed as:
 \begin{align}
 \label{equation:s_i}
 a_i &= \frac{\exp(\mathbf{f}_i^\top\mathbf{c}_{y_i})}{\sum\limits_{k=1}^C \exp(\mathbf{f}_i^\top\mathbf{c}_k)}.
 \end{align}
The softmax operation is used for normalising image's semantic relation over all classes.

In~\cite{goldberger2016training}, the softmax output (classification confidence/likelihood) is used to predict the true label. 
In this work, we apply it to estimate image's semantic relation to its label, i.e., the correctness degree of the label.

\subsection{Weighted Contrastive Loss} 
Now we have OSM for mining proper pairs and CAA for identifying outliers.
To demonstrate the effectiveness of OSM and CAA for deep metric learning, we integrate them into the traditional contrastive loss.
Specifically, first we generate OSM scores for both positive and negative pairs $(\mathbf{x}_i, \mathbf{x}_j) \in P \bigcup N$, and to deal with outliers at pair level we generate CAA scores for all pairs . That is,  
\begin{equation}
	\label{equation:w+ij2}
	w^+_{ij} = s^+_{ij}
	*
	 a_{ij},
\end{equation} 
\begin{equation}
	\label{equation:w-ij2}
	w^-_{ij} = s^-_{ij} * a_{ij},
\end{equation}
where $a_{ij}$ is the CAA score of pair $(\mathbf{x}_i, \mathbf{x}_j)$ and it is defined by $ \min(a_i, a_j)$. 

We integrate pair weights computed by Eq. (\ref{equation:w+ij2}) and Eq. (\ref{equation:w-ij2}) into the traditional contrastive loss to formulate our weighted contrastive loss:

\begin{equation}
	\label{equation:WCL_P}
	L_{WCL}(P) = \frac{1}{2}  \frac{\sum\limits_{(x_i, x_j) \in P} w^+_{ij}  {d_{ij}}^2}{\sum\limits_{(x_i, x_j) \in P} w^+_{ij}}
\end{equation} 
\begin{equation}
	\label{equation:WCL_N}
	L_{WCL}(N) = \frac{1}{2}  \frac{\sum\limits_{(x_i, x_j) \in N} w^-_{ij}  {\max(0, \alpha - d_{ij})}^2}{\sum\limits_{(x_i, x_j) \in N} w^-_{ij}}
\end{equation} 

\begin{equation}
	\begin{aligned}
		&L_{WCL}(P,N) = (1 - \lambda) L_{WCL}(P) + \lambda L_{WCL}(N)
	\end{aligned}
\end{equation}
where hyperparameter $\lambda$ controls the contribution of the positive set and the negative set towards final contrastive loss. 
In our experiments, we fix $\lambda = 0.5$, treating the positive set and negative set equally.
The denominators in Eq. (\ref{equation:WCL_P}) and Eq. (\ref{equation:WCL_N}) are used for normalisation.
See Figure~\ref{fig:overview} for the illustration of weigted contrastive loss with comparison to the traditional contrastive loss. 

 In particular, the number of negative pairs ($ck(ck-k)/2$) is much larger than the number of positive pairs ($ck(k-1)/2$). If we normalise all the dissimilar pairs and similar pairs together, the class imbalance problem makes learning unstable. Therefore, we apply independent normalisation for the positive samples $P$ and negative samples $N$. 

Following~\cite{sohn2016improved,tadmor2016learning}, we construct an unbiased estimate of the full gradient by relying on all possible $m(m-1)/2$ pairs in each mini-batch (batch size $m \ll$ data size). 
The layers for online pair construction and contrastive loss computation are put behind CNN embeddings, thus each image goes through CNN only once to obtain one embedding. Each embedding vector is used to construct multiple pairs with other embeddings and compute contrastive loss.  

\setlength{\tabcolsep}{3.1pt}
\begin{table*}[!t]
	\caption{Comparison with state-of-the-art methods on CARS196 and CUB-200-2011 in terms of Recall@\emph{K} (\%). The raw images are used for training and testing for methods in the first group. The cropped images are used for training and testing for methods in the second group. * indicates cascaded models are applied for sample mining and learning embeddings.}
	\label{literature_cub_cars}
	\centering
	\begin{tabular}{lcccccc|ccccccc}
		\hline
		& \multicolumn{6}{c|}{CARS196} & \multicolumn{6}{c}{CUB-200-2011} \\ 
		\cmidrule(r){2-13}
		\emph{K} & 1 & 2 & 4 & 8 & 16 & 32 & 1 & 2 & 4 & 8 & 16 & 32 \\
		\hline
		Contrastive \cite{bell2015learning} & 21.7 & 32.3 & 46.1 & 58.9 & 72.2 & 83.4 & 26.4 & 37.7 & 49.8 & 62.3 & 76.4 & 85.3 \\
		Triplet \cite{schroff2015facenet} & 39.1 & 50.4 & 63.3 & 74.5 & 84.1 & 89.8 & 36.1 & 48.6 & 59.3 & 70.0 & 80.2 & 88.4 \\
		LiftedStruct \cite{song2016deep} & 49.0 & 60.3 & 72.1 & 81.5 & 89.2 & 92.8 & 47.1 & 58.9 & 70.2 & 80.2 & 89.3 & 93.2 \\
		Binomial Deviance \cite{ustinova2016learning} & -- & -- & -- & -- & -- & -- & 52.8 & 64.4 & 74. 7 & 83.9 & 90.4 & 94. 3 \\
		Histogram Loss \cite{ustinova2016learning} & -- & -- & -- & -- & -- & -- & 50.3 & 61.9 & 72.6 & 82.4 & 88.8 & 93.7 \\
		
		Smart Mining \cite{harwood2017smart} & 64.7 & 76.2 & 84.2 & 90.2 & -- & -- & 49.8 & 62.3 & 74.1 & 83.3 & -- & -- \\
		
		HDC* \cite{yuan2017hard} & 73.7 & 83.2 & 89.5 & 93.8 & 96.7 & 98.4 & 53.6 & 65.7 & 77.0 & 85.6 & 91.5 & \textbf{95.5} \\
		Ours & \textbf{74.0} & \textbf{83.8} & \textbf{90.2} & \textbf{94.8}  & \textbf{97.3}  & \textbf{98.6}  & \textbf{55.3} & \textbf{67.3} & \textbf{77.5} & \textbf{85.8} & \textbf{91.8} & 95.4 \\
		\hline
		PDDM+Triplet \cite{huang2016local} & 46.4 & 58.2 & 70.3 & 80.1 & 88.6 & 92.6 & 50.9 & 62.1 & 73.2 & 82.5 & 91.1 & 94.4 \\
		PDDM+Quadruplet \cite{huang2016local} & 57.4 & 68.6 & 80.1 & 89.4 & 92.3 & 94.9 & 58.3 & 69.2 & 79.0 & 88.4 & 93.1 & 95.7 \\
		HDC* \cite{yuan2017hard} & 83.8 & 89.8 & 93.6 & 96.2 & 97.8 & 98.9 & 60.7 & 72.4 & 81.9 & 89.2 & 93.7 & 96.8 \\   
		Ours & \textbf{85.5} & \textbf{91.5} & \textbf{95.1} & \textbf{97.2} & \textbf{98.5} & \textbf{99.2} & \textbf{62.3} & \textbf{73.2} & \textbf{83.3} & \textbf{89.6} & \textbf{94.1} & \textbf{96.9} \\    
		\hline
	\end{tabular}
\end{table*}

\begin{table*}[!ht]
	\caption{Video-based person re-identification datasets: MARS and LPW. \# indicates the number of corresponding terms. DT failure denotes whether there exists detection or tracking failure in the tracklets. }
	\label{dataset_MARS_LPW}
	\centering
	\begin{tabular}{lcccccccc}
		\hline
		Datasets & \#identities & \#boxes & \#tracklets & \#cameras  & DT failure & resolution & annotation & evaluation \\
		\hline
		MARS & 1261 & 1,067,516 & 20,715 & 6 & Yes & 256x128 & detector & CMC \& mAP \\
		LPW & 2731 & 590,547 & 7694 & 11 & No & 256x128 & detector & CMC \\
		\hline      
	\end{tabular}
\end{table*}

\begin{table*}[!ht]
	\caption{Comparison with state-of-the-art methods on MARS in terms of CMC(\%) and mAP(\%). The methods using attention mechanism are indicated by `Attention' column. }
	\label{literature_mars}
	\centering
	\begin{tabular}{lccccc}
		\hline
		Methods & Attention & 1 & 5 &  20  & mAP \\
		\hline
		%
		IDE (ResNet50) \cite{zhong2017re} & No & 62.7 & -- &  -- & 44.1 \\
		IDE (ResNet50)+XQDA \cite{zhong2017re} & No & 70.5 & -- &  -- & 55.1 \\
		IDE (ResNet50)+XQDA+Re-ranking \cite{zhong2017re} & No & 73.9 & -- &  -- & 68.5 \\
		CNN+RNN \cite{mclaughlin2017video}        &  No & 43.0 & 61.0 &  73.0 & --\\
		CNN+RNN+XQDA \cite{mclaughlin2017video} & No & 52.0 & 67.0 &  77.0 & --\\
		AMOC+EpicFlow \cite{liu2017video} & No & 68.3 & 81.4 & 90.6 & 52.9 \\
		ASTPN \cite{xu2017jointly}          & Yes & 44.0  & 70.0    &  81.0 & -- \\
		SRM+TAM \cite{zhou2017see} & Yes & 70.6 & 90.0 & \textbf{97.6} & 50.7 \\
		RQEN \cite{song2017region} & Yes & 73.7 & 84.9 &  91.6 & 51.7 \\
		RQEN+XQDA+Re-ranking \cite{song2017region} & Yes & 77.8 & 88.8 & 94.3 & 71.1 \\
		DRSA \cite{li2018diversity} & Yes & 82.3 & -- & -- & 65.8\\
		CAE \cite{chen2018video} & Yes & 82.4 & 92.9 & -- & 67.5\\
		\hline
		Ours & Yes & \textbf{84.7} & \textbf{94.1} & 97.0 & \textbf{72.4} \\
		Ours + Re-ranking\cite{zhong2017re} & Yes & \textbf{86.0}  & \textbf{94.4} & 97.1 & \textbf{81.0} \\
		\hline  
	\end{tabular}
\end{table*}

\section{Experiments}

We compare our method with state-of-the-art embedding methods on two tasks from different domains, i.e., fine-grained visual categorisation and video-based person re-identification. For both tasks, testing and training classes are disjoint. This setting is common as the true deep metric should be class-agnostic.  

\paragraph{Implementation details.}  We use GoogleNet with batch normalisation \cite{ioffe2015batch} as the backbone architecture. 
%
For all datasets, the images are resized to $224 \times 224$ during training and testing. 
No data augmentation is applied for training and testing.
For fine-grained categorisation, we set $c=8$ and $k=7$ for each mini-batch, while $c=3$ and $k=18$ for video-based person re-identification.  
We set empirically ${\sigma_{\mbox {\tiny OSM}}} = 0.8$ and ${\sigma_{\mbox {\tiny CAA}}} = 0.18$ for all experiments to avoid adjusting them based on the specific dataset, making it general and fair for comparision. 
When training a model, the weights are initialised by the pretrained model on ImageNet \cite{russakovsky2015imagenet}. 
For optimisation, Stochastic Gradient Decent (SGD) is used with a learning rate of 0.001, a momentum of 0.9. 
%
%
The margin of weighted contrastive loss $\alpha$ is set to 1.2 for all experiments. 
We implement our method in Caffe \cite{jia2014caffe} deep learning framework. We will release the source code and trained models for the ease of reproduction. 


\subsection{Fine-Grained Visual Categorisation}
Fine-grained visual categorisation task is widely used for evaluating deep metric learning methods. 

\noindent
\textbf{Datasets and evaluation protocol.}
We conduct experiments on two benchmarks:
\begin{itemize}
	\item CUB-200-2011 contains 11,788 images of 200 species of birds. The first 100 classes (5,864 images) are used for training and the remaining classes (5,924 images) for testing.
	\item CARS196 is composed of 16,185 images of 196 types of cars. The first 98 classes containing 8,054 images are for training and the remaining 98 classes (8,131 images) for testing.
\end{itemize}
Both datasets come with two versions: raw images (containing large background with respect to a region of interest) and cropped images (containing only regions of interest). For fair comparison, we follow standard evaluation protocol \cite{song2016deep} and report the retrieval results using Recall@\textit{K} metric. 
For each query, the Recall@\textit{K} is 1 if at least one positive in the gallery appears in its top \textit{K} nearest neighbours. During testing, every image is served as a probe in turn and all the other images compose the gallery set. \\
\noindent
\textbf{Comparative evaluation.}
We compare our method with state-of-the-art methods in Table~\ref{literature_cub_cars}. All the compared methods use GoogleNet as the base architecture. We have the following observations: 
\begin{itemize}
	\item In both settings of datasets, i.e., raw images and cropped images, our method outperforms all the compared methods. Generally, our method improves state-of-the-art performance by around $1.5\%$.
	\item The only method that is competitive to ours is HDC~\cite{yuan2017hard}. However, it is important to note that HDC~\cite{yuan2017hard} is based on cascaded embedding requiring several models, while ours is a single model.  
\end{itemize}

\begin{table}[!t]
	\caption{Comparison with state-of-the-art methods on LPW in terms of CMC (\%). The methods using attention mechanism are indicated by `Attention' column.}
	\label{literature_lpw}
	\centering
	\begin{tabular}{lccccc}
		\hline
		Methods & Attention & 1 & 5  & 20\\
		\hline
		GoogleNet\cite{song2017region}  & No & 41.5  & 66.7 & 86.2  \\
		RQEN\cite{song2017region}  & Yes & 57.1 & 81.3 & 91.5 \\ 
		\hline
		Ours & Yes & \textbf{71.7} & \textbf{89.8} & \textbf{96.6} \\
		\hline      
	\end{tabular}
\end{table}

\begin{table*}[!ht]
	\caption{Ablation studies on CUB-200-2011, LPW, MARS in terms of Recall@\textit{K} (\%) and mAP (\%).}
	\label{ablation_study}
	\centering
	\begin{tabular}{lcccccc|ccc|cccc}
		\hline
		&  \multicolumn{6}{c}{CUB-200-2011} & \multicolumn{3}{|c}{LPW} & \multicolumn{4}{|c}{MARS} \\ 
		\cmidrule(r){2-14}
		\emph{K} & 1 & 2 & 4 & 8 & 16 & 32 & 1 & 5 & 20 & 1 & 5 & 20 & mAP \\
		\hline
		GoogleNet & 47.8 & 61.1 & 73.2 & 83.2 & 90.4 & 95.4  & -- & -- & -- & -- & -- & -- & -- \\
		\hline
		Baseline & 52.0 & 65.0 & 76.2 & 84.8 & 90.9 & 95.0  & 61.9 & 84.4 & 94.3  & 79.1 & 91.5 & 95.9 & 66.6 \\
		
		OSM & 54.2 & 66.8 & 76.9 & 85.9 & 91.6 & 95.5  & 69.3 & 89.8 & 95.4    & 82.3 & 93.1 & 96.0 & 70.5 \\
		
		OSM+CAA & \textbf{55.3} & \textbf{67.3} & \textbf{77.5} & \textbf{85.8} & \textbf{91.8} & \textbf{95.4}  & \textbf{71.7} & \textbf{89.8} & \textbf{96.6} & \textbf{84.7} & \textbf{94.1} & \textbf{97.0} & \textbf{72.4} \\
		
		\hline
	\end{tabular}
\end{table*}

\subsection{Video-based Person Re-identification}

To further evaluate the effectiveness of our method, we conduct experiments on video-based person re-identification task.

\noindent
\textbf{Datasets and evaluation protocol.}
In this task, we evaluate our methods on two large-scale benchmark datasets: 
MARS \cite{zheng2016mars} consists of images (frames) with huge variations due to camera setup, yielding many noisy images and outliers, and LPW \cite{song2017region} is a cross-scene video dataset. Note that LPW is more challenging than MARS dataset, since data is collected under different scenes, whereas MARS is collected in only one scene, around university campus. The detailed information about MARS and LPW datasets are summarised in Table \ref{dataset_MARS_LPW}. We follow exactly the evaluation setting in \cite{zheng2016mars} and \cite{song2017region}, respectively. 
We report the Cumulated Matching Characteristics (CMC)\footnote{CMC@\textit{K} is the same as Recall@\textit{K}. Generally, the term CMC is used in person re-identification.} results for both datasets. We also compute the mean average precision (mAP) for MARS. 

\noindent
\textbf{Comparative evaluation.}
The results of MARS and LPW are shown in Table~\ref{literature_mars} and Table~\ref{literature_lpw} respectively. The results on both datasets demonstrate the effectiveness of our proposed approach. Our observations are as follows: 
\begin{itemize}
	\item On MARS dataset, our method outperforms all the methods by a large margin. Compared with CAE~\cite{chen2018video}\footnote{For CAE, we present the results of complete sequence instead
	of multiple snippets so that it can be compared with other methods.
	Multiple snippets can be regarded as data augmentation in the test phase.},  ours improves 2.3\% and 4.9\% in terms of CMC-1 and mAP respectively; When using re-ranking as post-processing in the test phase, the performance of our method further increases.
	\item On LPW dataset, we can see that our method again outperforms state-of-the-art methods by a large margin. 
\end{itemize}

\subsection{Ablation Study}

Our key contributions in this paper are OSM and CAA. To evaluate the contribution of each module, we conduct ablation studies on CUB-200-2011, LPW, and MARS datasets. The details are as follows:
\begin{itemize}
	\item Baseline: Neither OSM or CAA is used. Online pair construction is applied. The weight of each pair is fixed as one in the loss functions Eq.~(\ref{equation:WCL_P}) and Eq.~(\ref{equation:WCL_N}). 
	Specifically, $w^+_{ij} = 1$ and $w^-_{ij} = 1$.
	\item OSM: When mining negatives, OSM generates higher scores for difficult negatives and lower scores for trivial negative samples.
	When mining positives, OSM generates higher scores for local positives to preserves intraclass variances. CAA is not used. Specifically, $w^+_{ij} = s^+_{ij}$ and $w^-_{ij} = s^-_{ij}$.
	
	\item OSM+CAA: As mentioned before, OSM is prone to outliers similar to other mining methods, which is an inherent problem of mining approaches. Therefore, we combine OSM with CAA to conduct soft mining as well as address outliers. Specifically, $w^+_{ij} = s^+_{ij} * a_{ij}$ and $w^-_{ij} = s^-_{ij} * a_{ij}$.
\end{itemize}


The results are summarized in Table~\ref{ablation_study}. We can see that OSM alone can achieve very competitive performance compared with the baseline. After integrating CCA to OSM, the accuracy increases on all datasets by about 1-2\%. It is easy to see that both OSM and CCA contribute positively toward the final performance. 

Furthermore, we notice that  
test set classes of CUB-200-2011 overlaps with ImageNet \cite{russakovsky2015imagenet} dataset. To verify how much actually we improve, we show the performance of pre-trained GoogleNet \cite{ioffe2015batch} without any fine-tuning. The result is shown in the first row of Table \ref{ablation_study}. Just using GoogleNet with pre-trained weights, the performance is 47.8\% -- already beating most of the methods shown in Table ~\ref{literature_cub_cars}. In contrast, our method outperforms pre-trained GoogleNet by a large margin.




\section{Conclusion}
\label{conclusion}

In this work, we propose a simple yet effective mining method named OSM. Importantly, OSM not only mines nontrivial sample pairs for accelerating convergence, but also learns extended manifolds for preserving intraclass variances. Besides, we propose CAA to address the impact of outliers. Finally, to evaluate the effectiveness of these two modules in action, we propose weighted contrastive loss by combining OSM and CAA to learn discriminative embeddings. The experiments on fine-grained visual categorisation and video-based person re-identification with four datasets demonstrate the superiority of our method. For future work, it is interesting to see the integration of OSM and CAA for other kinds of losses such as triplet loss.  

\bibliography{AAAI-WangX.3625}
\bibliographystyle{aaai}
\end{document}